\documentclass{article}

\PassOptionsToPackage{numbers, compress}{natbib}
\usepackage[preprint]{neurips_2026}

\usepackage[utf8]{inputenc} 
\usepackage[T1]{fontenc}    
\usepackage[hypertexnames=false]{hyperref}       
\usepackage{url}            
\usepackage{graphicx}       
\usepackage{wrapfig}        
\usepackage{booktabs}       
\usepackage{amsfonts}       
\usepackage{amsmath}        
\usepackage{nicefrac}       
\usepackage{microtype}      
\usepackage[table]{xcolor}  
\usepackage{enumitem}       
\usepackage{placeins}       
\usepackage{float}         
\usepackage{xcolor}
\usepackage{xspace}
\definecolor{stormcolor}{RGB}{45,55,90}
\definecolor{ModelPurple}{HTML}{E1D6FF}
\definecolor{ModelGreen}{HTML}{E6F4EA}
\definecolor{ToolThinkBar}{HTML}{FDECC8}
\definecolor{VideoThinkBar}{HTML}{DCEBFF}
\definecolor{YiyangOrange}{HTML}{F59E0B}
\definecolor{AndongGreen}{HTML}{61F527}
\definecolor{YimingPurple}{HTML}{682699}
\definecolor{YixiaoBlue}{HTML}{1E90FF}

\newcommand{\stormtitlelogo}{%
\raisebox{-0.22em}[0pt][0pt]{%
\includegraphics[height=1.35em]{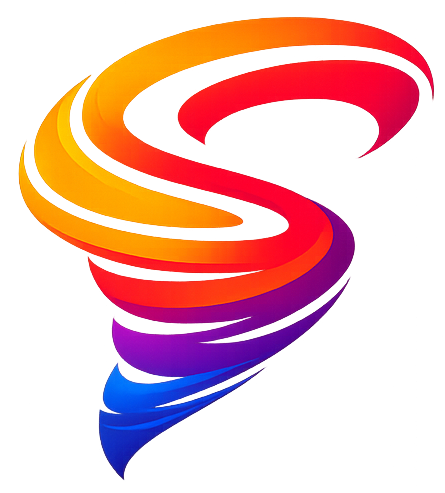}}%
}

\newcommand{\storminlinelogo}{%
\raisebox{-0.14em}[0pt][0pt]{%
\includegraphics[height=1.0em]{logo.png}}%
}

\DeclareRobustCommand{\storms}{%
\texorpdfstring{%
\mbox{%
\storminlinelogo
\hspace{-0.12em}%
{\color{stormcolor}\textbf{\textit{T\kern-0.05em O\kern-0.04em R\kern-0.03em M}}}}%
}{STORM}%
\xspace
}

\DeclareRobustCommand{\STORM}{%
\hspace{-0.45em}%
\stormtitlelogo
\hspace{-0.18em}%
{\color{stormcolor}
\fontsize{18}{18}\selectfont
\textit{T\kern-0.05em O\kern-0.04em R\kern-0.03em M}}%
}

\newcommand{\stormpdftitle}{STORM: Internalized Modeling for Spatial-Temporal Reasoning in Video-Language Models}

\title{\texorpdfstring{\STORM}{STORM}: Internalized Modeling for Spatial-Temporal Reasoning in Video-Language Models}
\hypersetup{pdftitle={\stormpdftitle}}

\author{%
  \normalfont
  \textbf{Yiming Liang}\textsuperscript{1}\thanks{Equal contribution.} \quad
  \textbf{Yixiao Chen}\textsuperscript{2}\footnotemark[1] \quad
  \textbf{Yiyang Zhou}\textsuperscript{3}\thanks{Project lead.} \quad
  \textbf{Yixuan Wang}\textsuperscript{3} \quad
  \textbf{Shoubin Yu}\textsuperscript{3}\\
  \textbf{Andong Deng}\textsuperscript{4} \quad
  \textbf{Fuxiao Liu}\textsuperscript{5} \quad
  \textbf{Qin Zhang}\textsuperscript{6} \quad
  \textbf{Chen Chen}\textsuperscript{4} \quad
  \textbf{Mohit Bansal}\textsuperscript{3} \quad
  \textbf{Huaxiu Yao}\textsuperscript{3}\\[0.35em]
  \textsuperscript{1}Purdue \quad
  \textsuperscript{2}Harvard \quad
  \textsuperscript{3}UNC \quad
  \textsuperscript{4}UCF \quad
  \textsuperscript{5}NVIDIA \quad
  \textsuperscript{6}Physion Labs
}

\begin{document}

\maketitle

\begin{abstract}
Many video reasoning tasks require tracking motion, temporal order, and evolving visual states across frames. Existing methods built on large vision--language models (LVLMs) often address this challenge by externalizing reasoning through textual chain-of-thought (CoT), keyframe selection, repeated frame reinsertion, or external tool use. While effective, such pipelines increase inference-time latency and engineering complexity, and they force temporal-visual evidence to be serialized into text or repeatedly re-encoded from frames. Inspired by the intuition that visual reasoning can occur implicitly before verbalization, we propose \storms (\textbf{S}patial-\textbf{T}emporal reas\textbf{O}ning via inte\textbf{R}nalized \textbf{M}odeling), a two-stage framework that teaches LVLMs to reason through bounded continuous latent trajectories instead of explicit textual CoT. In Stage~I, \storms aligns latent tokens with thought-video representations derived from generated videos, grounding the latent states in dynamic visual evidence. In Stage~II, the model is further trained with answer-only supervision, encouraging the reasoning process to be internalized without step-by-step annotations. Generated thought videos are used only during training; at inference, \storms performs a bounded latent rollout without regenerating videos, reinserting frames, or invoking external visual tools. Experiments on VideoMME, MVBench, TempCompass, and MMVU show that \storms improves video reasoning accuracy while substantially reducing inference overhead compared with tool or video-generation-based reasoning pipelines. Code is availabe at https://github.com/aiming-lab/storm
\end{abstract}

\section{Introduction}
\label{sec:intro}

Large vision--language models (LVLMs) have achieved strong performance on image understanding by aligning visual representations with large language models (LLMs)~\citep{Bai2025Qwen25VLTR,li2024llava}. However, extending this success to video reasoning remains challenging. Unlike static images, video understanding requires tracking motion, temporal order, and evolving scene states across frames. Although textual chain-of-thought (CoT) reasoning can help structure the reasoning process, it often struggles to faithfully represent fine-grained temporal and visual evidence. Video reasoning inherently involves multiple interacting factors, such as object dynamics, spatial relationships, and scene transitions, that evolve simultaneously over time, whereas textual rationales must linearize these entangled signals into a discrete token sequence.

To address this limitation, existing approaches often externalize reasoning by repeatedly revisiting visual inputs through frame reinsertion, temporal decomposition, or external tool-based pipelines~\citep{arnab2026temporal,li2026universal,wang2026time,zhou2026reagent}. These methods improve reasoning performance by exposing intermediate visual evidence during inference, but they also introduce substantial latency, engineering complexity, and sensitivity to retrieval or tool failures. This reveals a fundamental trade-off in video reasoning: strong reasoning performance benefits from preserving rich intermediate visual-temporal information, yet explicitly generating, retrieving, or reprocessing such evidence at inference time can significantly reduce efficiency and robustness.

These limitations motivate reasoning directly in continuous latent space, where hidden representations can compactly preserve visual-temporal information without repeatedly materializing it into explicit textual or visual intermediates. Unlike discrete textual reasoning, latent representations naturally support superposed and distributed features, allowing multiple reasoning factors to coexist and interact within the same hidden states~\citep{zhu2025emergence}. Building on this insight, for complex video reasoning, we ask whether intermediate visual evidence can be used during training to induce an internal latent reasoning process, while keeping inference as a single bounded decoding procedure. Rather than explicitly revisiting visual evidence at test time, the model learns to internalize temporal-visual reasoning within hidden states.

\begin{figure*}[t]
    \centering
    \includegraphics[width=0.98\textwidth]{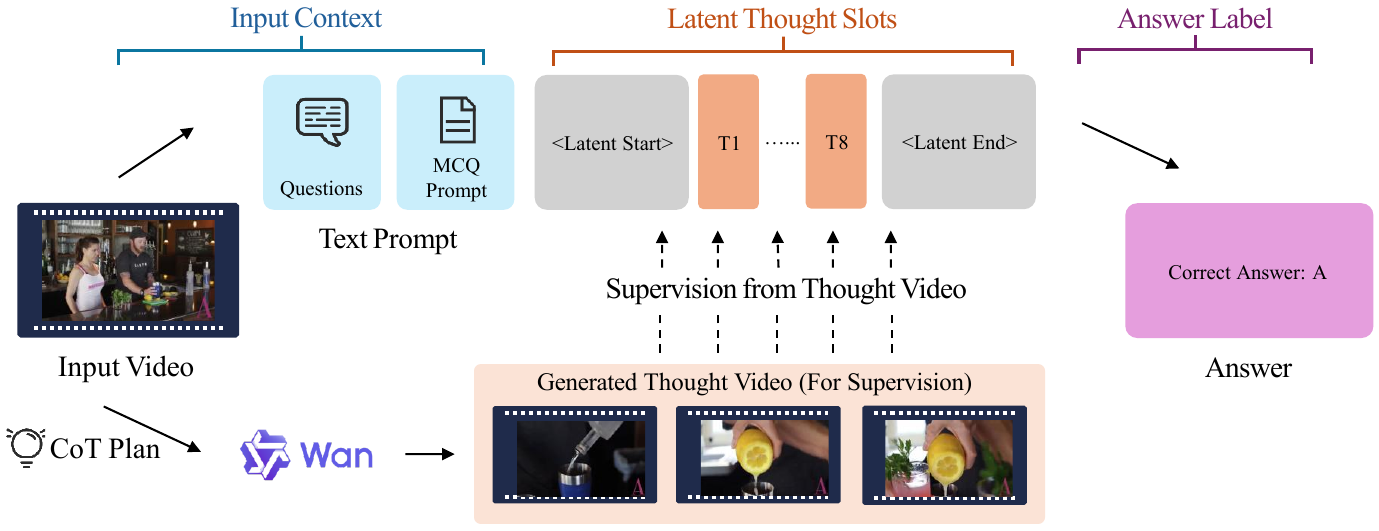}
    \caption{Interleaved \storms training sequence. A generated thought video provides training-time dynamic supervision for the latent tokens, encouraging them to encode question-relevant temporal evidence before the model generates the final answer.} 
    \label{fig:interleaved_training_data}
    \vspace{-3mm}
\end{figure*}

In this paper, we introduce \storms, a two-stage framework that teaches a video-language model to reason through continuous latent rather than explicit textual CoT or tools. The key idea is to expose the model to rich dynamic evidence during training, but to require it to use only compact hidden-state rollouts at inference time.
Specifically, as shown in Figure~\ref{fig:interleaved_training_data}, each training instance contains three aligned signals: (1) sampled keyframes from the source video, (2) a generated thought video that visualizes question-relevant motion and state changes, and (3) the target answer sequence. We first construct thought-video supervision by combining selected keyframes with a teacher reasoning plan, producing a short dynamic trace that externalizes the visual evidence needed for the answer. We then insert a small number of latent tokens between the video-question context and the final answer, and supervise these latent states to encode thought-video features before answer generation.

\begin{wrapfigure}[15]{r}{0.49\textwidth}
    \vspace{-8pt}
    \centering
    \includegraphics[width=0.9\linewidth]{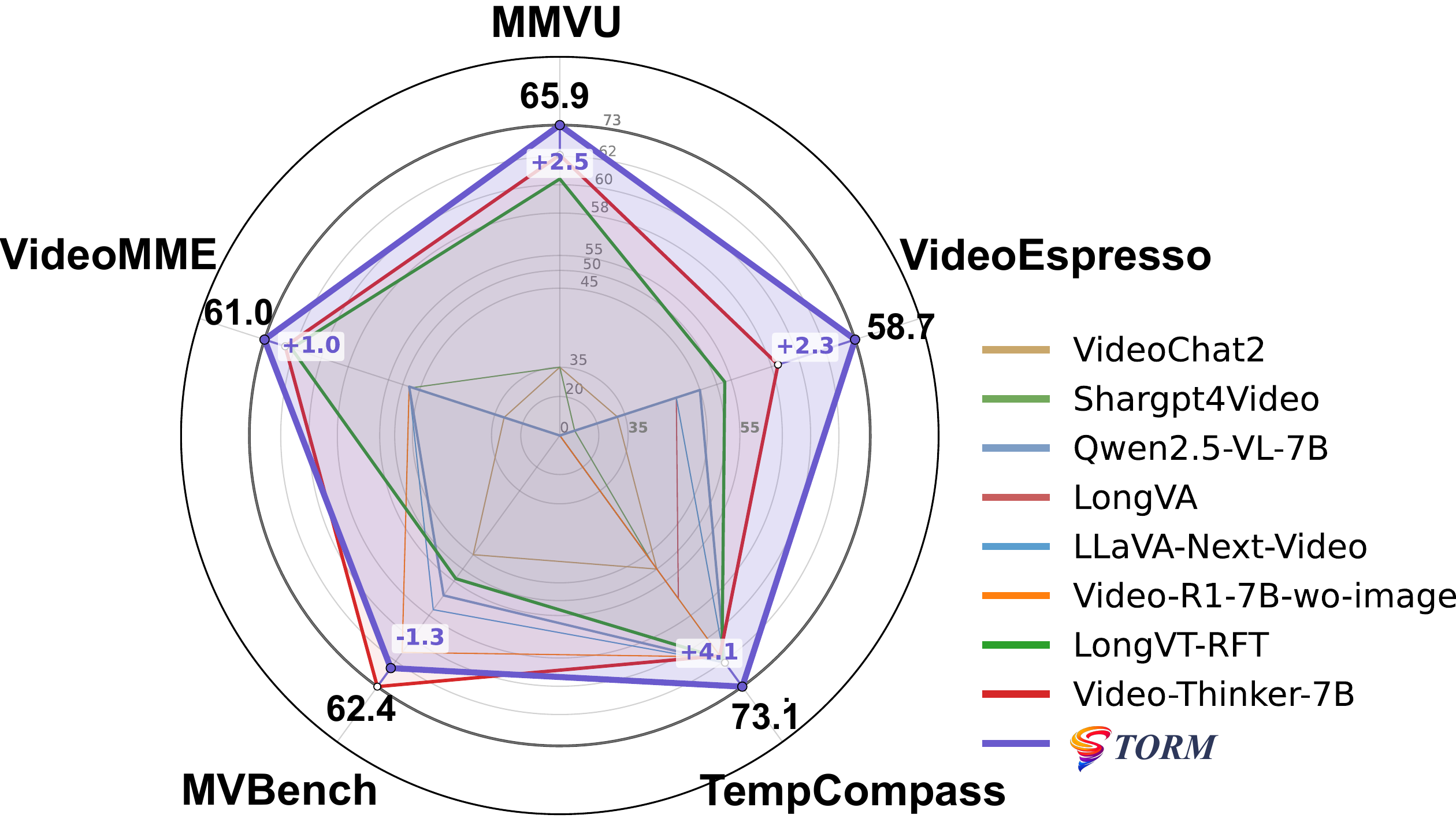}
    \caption{Comparison between \storms and representative video reasoning methods across benchmarks.}
    \vspace{-5mm}
    \label{fig:radar}
\end{wrapfigure}

The training procedure is organized in two stages. In \textbf{Stage~I}, we jointly optimize answer-token loss and latent-state alignment to thought-video representations, grounding the latent slots in temporally relevant visual evidence without requiring textual CoT supervision. In \textbf{Stage~II}, we adopt a Coconut-like setting~\citep{hao2024training}: the latent slots remain in the sequence, but supervision is applied only to the final answer. This encourages the model to consolidate the grounded latent behavior into an answer-focused internal reasoning process.
At inference time, \storms does not generate thought videos, revisit frames, or call external visual tools. Once the model enters the latent segment, it performs a bounded continuous rollout through hidden states and then resumes normal text decoding for the final answer. In other words, dynamic visual evidence is used only to teach the latent states during training; at inference, the model reuses this learned behavior through a compact, fixed-budget latent trajectory without generating new videos via heavy external tools.

This design differs from recent ``thinking with images'' and latent-reasoning methods~\citep{xu2025visual,wu2024vsp,fu2024blink,ray2024sat,yang2025machine,li2025latent}. Prior approaches are often designed for static images, continuous text-space reasoning, explicit visual generation, or inference-time tool use. In contrast, \storms targets video reasoning directly by using generated videos only as training-time supervision and by internalizing spatial-temporal reasoning into bounded latent states. Figure~\ref{fig:radar} and our benchmark results show that this formulation improves video reasoning accuracy while substantially reducing the inference overhead associated with external reasoning pipelines. In summary, our main contributions are:
\begin{itemize}[leftmargin=*, itemsep=2pt, topsep=2pt]
    \item We introduce \storms, a video reasoning framework that performs internal dynamic simulation in continuous latent space instead of relying on explicit textual CoT or tool-heavy inference.
    \item We propose a two-stage training strategy: Stage~I aligns latent tokens with thought-video supervision plus answer-token loss, and Stage~II uses answer-only supervision to strengthen internalized reasoning.
    \item We build a thought-video supervision pipeline that aligns generated dynamic traces, keyframes, and QA annotations, enabling scalable training for latent video reasoning.
    \item We show that \storms improves video reasoning performance across benchmarks while reducing dependence on computationally expensive and operationally cumbersome test-time pipelines.
\end{itemize}

\section{Related Work}
\label{gen_inst}

\subsection{Latent Visual and Implicit Reasoning}
Recent work explores replacing explicit reasoning traces with compact continuous or latent representations to improve efficiency. Latent-visual methods augment VLM decoding with internal visual tokens or reconstruct task-relevant visual features, enabling multi-step visual inference without generating full-resolution images~\citep{yang2025machine, li2025latent}. Related multimodal latent-thinking methods further introduce modality-agnostic latent tokens or active visual imagination for spatial reasoning~\citep{ray2025mull, cao2025spatialdreamer}. In parallel, continuous-thought methods train language models to reason by feeding back hidden states or distilling textual CoT into shorter latent trajectories that preserve reasoning ability while reducing token overhead~\citep{hao2024training, deng2024explicit, shen2025codi}, while perception-token methods encode structured visual cues as dedicated latent inputs for fine-grained visual reasoning~\citep{bigverdi2025perception, yang2024large}. However, these methods are mostly designed for text reasoning, static images, or spatial imagination, and do not explicitly address long-horizon video reasoning with unified, temporally grounded, task-aware latent representations.

\subsection{Video--Language Models}
Recent video LVLMs extend image-based multimodal models to temporal understanding by improving video representation, video instruction tuning, and token-efficient temporal modeling. Generalist models such as Qwen2.5-VL~\citep{Bai2025Qwen25VLTR} and LLaVA-OneVision~\citep{li2024llava} unify image, multi-image, and video inputs to support broad multimodal instruction following. Video-specialized models such as LLaVA-Video~\citep{zhang2024llava} and VideoLLaMA3~\citep{zhang2025videollama} further strengthen video understanding through large-scale video instruction tuning and vision-centric multimodal training. In parallel, InternVL3~\citep{zhu2025internvl3}, InternVL3.5~\citep{wang2025internvl3}, SlowFast-LLaVA-1.5~\citep{xu2025slowfast}, and Eagle~2.5~\citep{chen2026eagle} explore stronger multimodal pretraining, long-context post-training, and token-efficient long-form video comprehension. These advances improve how videos are encoded and consumed by LVLMs, but they typically still perform reasoning over sampled or compressed visual tokens at inference time. In contrast, \storms focuses on internalizing dynamic spatial-temporal reasoning into bounded latent trajectories.

\subsection{Test-time Visual Reasoning}
A line of work improves multimodal reasoning by making intermediate visual evidence explicit during reasoning. At the image or multi-image level, methods such as CMMCoT and ReFocus use multimodal CoT, memory augmentation, or visual editing to support cross-image comparison and structured image understanding~\citep{zhang2026cmmcot, fu2025refocus}. For video understanding, frame-aware reasoning methods construct or train on reasoning traces that explicitly identify question-relevant frames, while temporal CoT methods iteratively select useful video context at inference time~\citep{ghazanfari2025chain, arnab2026temporal}. In parallel, tool-driven and active-perception frameworks call vision experts, zooming modules, selective revisitation mechanisms, or world models to gather additional visual evidence during reasoning~\citep{yang2023mm, shen2025zoomeye, chung2025don, zhang2025chain, yu2026and}. Related training-time approaches such as VisionCoach~\citep{lee2026visioncoach} show that visual prompting guidance can be internalized through reinforcement learning, allowing the model to remain prompting-free at inference time. In contrast, \storms uses generated thought videos as training-time supervision for bounded latent trajectories, avoiding repeated frame selection, visual editing, and external tool calls at inference while still capturing temporal structure and critical visual cues.

\section{\storms}
\label{headings}
We introduce \storms, a two-stage training framework that teaches an LVLM to reason through internal latent trajectories instead of explicit textual CoT. \storms uses three aligned signals for each QA pair: (1) sampled keyframes from the source video, (2) a generated thought video that externalizes dynamic reasoning evidence, and (3) the target answer token sequence.

\subsection{Overview}

\begin{figure*}[t]
    \centering
    \includegraphics[width=0.95\textwidth]{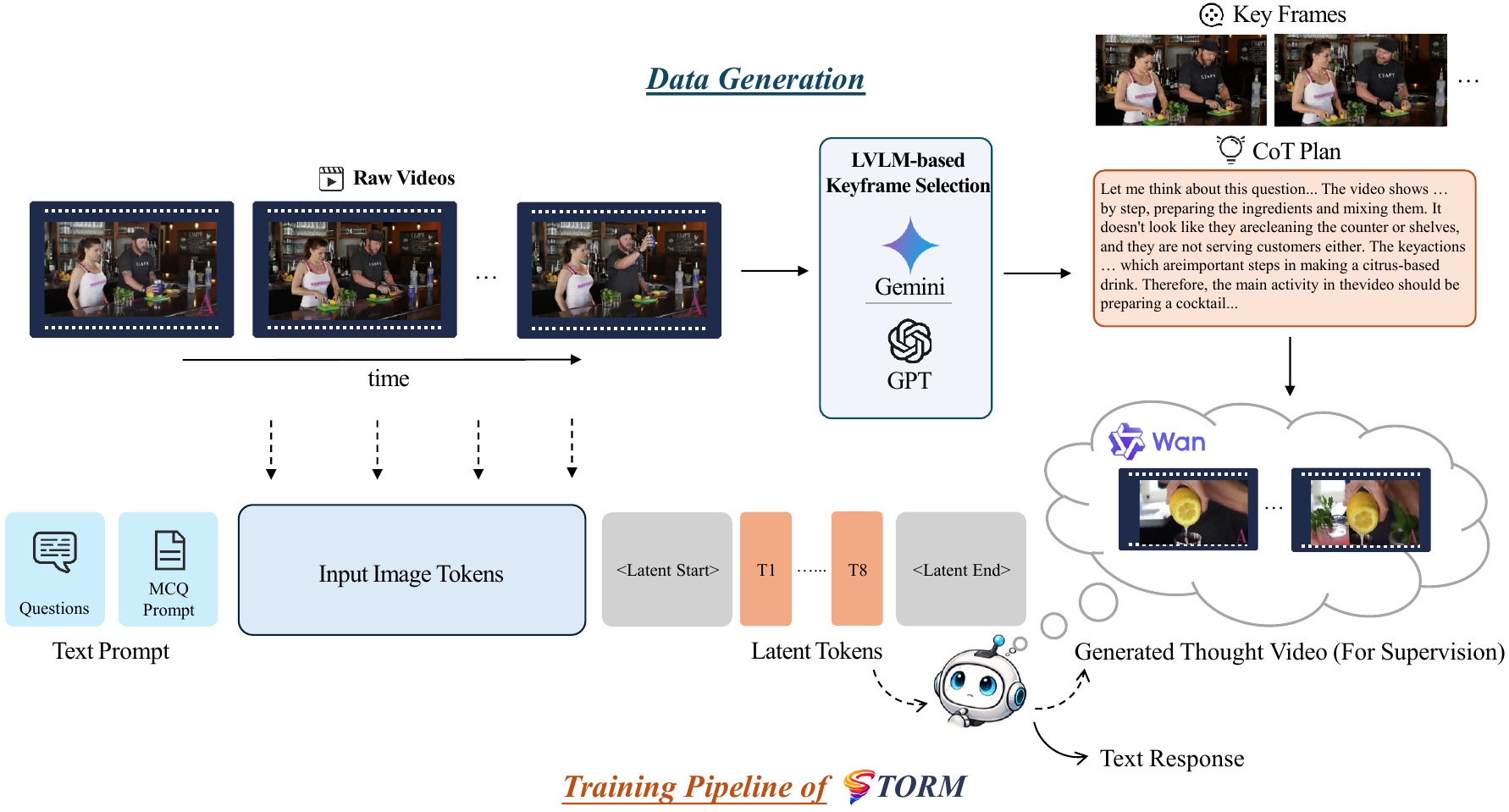}
    \caption{The overall workflow of the \storms framework, including keyframe input, CoT planning, thought-video generation for training-time supervision, latent simulation, and final answer prediction.}
    \label{fig:system_workflow}
\end{figure*}

Figure~\ref{fig:system_workflow} summarizes the overall \storms system workflow used in our method, given a video $\mathcal{V}=\{f_1,\ldots,f_T\}$ and question token sequence $\mathcal{Q}=\{q_1,\ldots,q_J\}$, the model predicts answer tokens $\mathcal{Y}=\{y_1,\ldots,y_N\}$:
\begin{equation}
P_\theta(\mathcal{Y}\mid\mathcal{V},\mathcal{Q})=
\prod_{t=1}^{N}P_\theta(y_t\mid y_{<t},\mathcal{V},\mathcal{Q}).
\label{eq:mvs_autoregressive}
\end{equation}

\storms  augments this process with a short latent simulation segment containing $K$ latent tokens (we use $K=8$). During this segment, the decoder hidden states are treated as simulation states rather than decoded text. Let $\mathbf{z}_i$ be the hidden state at latent position $i$. Answer generation is then conditioned on both the original video-question context and the latent simulation states:
\begin{equation}
P_\theta(\mathcal{Y}\mid\mathcal{V},\mathcal{Q},\mathcal{Z})=
\prod_{t=1}^{N}P_\theta(y_t\mid y_{<t},\mathcal{V},\mathcal{Q},\mathcal{Z}),
\quad
\mathcal{Z}=\{\mathbf{z}_1,\ldots,\mathbf{z}_K\}.
\label{eq:mvs_conditioned_answer}
\end{equation}

Intuitively, $\mathcal{Z}$ acts as an internal ``thought video'' summary: it stores dynamic visual evidence in compact latent form before answer emission.

During inference, the model follows the same latent-slot pattern autoregressively. Once the special token \texttt{<|latent\_start|>} is generated, the decoder enters a latent simulation mode. In this mode, subsequent discrete token predictions are replaced by \texttt{<|latent\_pad|>} placeholders in the token sequence, while the corresponding hidden states are fed back as continuous input embeddings for the next decoding step. The model remains in this mode until it generates \texttt{<|latent\_end|>}, after which decoding resumes in the normal text space for final answer generation. To bound the computation, we impose a maximum latent budget of $K$ slots, with $K=8$ in all main experiments. If the model exceeds this budget without producing \texttt{<|latent\_end|>}, we force the latent segment to terminate by inserting \texttt{<|latent\_end|>}, ensuring a fixed upper simulation length and stable inference latency.

\subsection{Data Pipeline}
\label{data_pipeline}
As illustrated in Figure~\ref{fig:system_workflow}, starting from video-QA samples, we construct interleaved training sequences for latent simulation supervision.

\textbf{Step 1: Keyframe extraction.}
For each sample, we use GPT-4o \citep{hurst2024gpt} to extract question-relevant keyframes from the source video and denote them as
$\mathcal{K}=\{k_1,\ldots,k_M\}$, where $M$ is the number of selected keyframes.

\textbf{Step 2: Thought-video construction.}
We denote the teacher-generated CoT sequence as $\mathcal{C}^{\mathrm{teach}}=\{c_1,\ldots,c_L\}$, where $L$ is the CoT length.
Conditioned on the first keyframe $k_1 \in \mathcal{K}$ and teacher-generated CoT $\mathcal{C}^{\mathrm{teach}}$, we generate thought videos using Wan 2.1 \citep{wan2025wan} with a conditional diffusion process.

\textbf{Step 3: Token alignment.}
As shown in figure~\ref{fig:interleaved_training_data}, we align three spans in one sequence: keyframe visual tokens, latent simulation slots, and answer tokens. This alignment yields direct supervision for latent simulation (on latent spans) and answer generation (on text spans), while avoiding explicit textual CoT supervision.

\subsection{Two-Stage Training}

\storms uses two training stages to first ground latent simulation, then internalize reasoning behavior.

\paragraph{Stage~I: Latent-grounded supervised training.}
In Stage~I, latent slots are supervised to match thought-video features, while answer tokens are trained with standard language modeling loss.

{\setlength{\abovedisplayskip}{4pt}
\setlength{\belowdisplayskip}{4pt}
\setlength{\abovedisplayshortskip}{2pt}
\setlength{\belowdisplayshortskip}{2pt}
Let $\mathbf{z}_i$ denote the predicted latent state at slot $i$, and let $\mathbf{g}_i$ denote the corresponding thought-video target feature. For each training sample, we first encode the generated thought video with the Qwen visual encoder and obtain frame-level visual tokens. We concatenate the tokens from the sampled thought-video frames along the token dimension, producing a sequence $\mathbf{H} \in \mathbb{R}^{M \times d}$, where $M$ is the total number of visual tokens and $d$ is the hidden dimension. Since $M$ may vary across samples, we apply one-dimensional adaptive average pooling over the token dimension to compress $\mathbf{H}$ into a fixed-length sequence of $K$ target vectors, $\mathbf{G} = \{\mathbf{g}_i\}_{i=1}^{K} \in \mathbb{R}^{K \times d}$. Each $\mathbf{g}_i$ is therefore the average representation of one adaptive segment of the thought-video token sequence, providing compact temporal-visual supervision for the corresponding latent slot $\mathbf{z}_i$. We then use an MSE objective to align each predicted latent state $\mathbf{z}_i$ with its pooled target $\mathbf{g}_i$. Stage~I losses are:

\begin{equation}
\mathcal{L}_{\mathrm{latent}}=\frac{1}{K}\sum_{i=1}^{K}\left\lVert \mathbf{z}_i-\mathbf{g}_i \right\rVert_2^2,
\qquad
\mathcal{L}_{\mathrm{ans}}=-\sum_{t=1}^{N}\log P_\theta(y_t\mid y_{<t},\mathcal{V},\mathcal{Q},\mathcal{Z}).
\label{eq:mvs_stage1_losses}
\end{equation}
The Stage~I objective is
\begin{equation}
\mathcal{L}_{\mathrm{stage1}}=\mathcal{L}_{\mathrm{ans}}+\lambda\mathcal{L}_{\mathrm{latent}},
\label{eq:mvs_stage1_obj}
\end{equation}
where $\lambda$ controls latent alignment strength. In other words, Eq.~\ref{eq:mvs_stage1_losses} trains the model to compress the pooled visual dynamics of the thought video into a short latent simulation trajectory, while the answer-token loss ensures that these simulated states remain useful for downstream prediction.}

\paragraph{Stage~II: Answer-only consolidation.}
In Stage~II, we remove direct latent supervision and optimize only $\mathcal{L}_{\mathrm{ans}}$. Latent slots are still present in the input pattern, but the model receives no explicit loss on latent states. This forces latent reasoning to become self-organized and robust, while preserving the answer-focused objective used at inference.

Overall, this two-stage design trains the model to first learn meaningful thought-video-aligned latent dynamics, then rely on those internal dynamics without handcrafted latent targets.

\section{Experiment}
\label{exp}

This section provides a complete empirical evaluation of \storms. We first summarize the benchmark suite used to assess general video understanding and reasoning-heavy video QA, then describe the training data, backbone, optimization setup, latent-token budget, and hardware configuration. We next report the main results across general and reasoning-heavy benchmarks, including a smaller-backbone scalability study and an inference-latency comparison against tool- and video-generation-based reasoning pipelines. Finally, we analyze the design choices behind \storms through ablations on latent size and two-stage training, visualize latent-token alignment, and probe whether the learned latent states preserve video-specific information through same-video retrieval diagnostics.

\begin{table*}[t]\small
\centering
\caption{General video benchmark results. The orange/blue bars indicate think-with-tools (LongVT) and think-with-video (Video-Thinker) with 32 max frames for a single round. Except for those two methods, all experiments use uniform sampling with 32 frames.}
\label{tab:general_benchmarks}
\setlength{\tabcolsep}{4pt}
\resizebox{\textwidth}{!}{%
\begin{tabular}{lcccc}
\toprule
\textbf{Model} & \textbf{VideoMME} & \textbf{MVBench} & \textbf{TempCompass} & \textbf{Frames} \\
\midrule
\rowcolor{VideoThinkBar}\multicolumn{5}{l}{\textit{Proprietary Models}} \\
\midrule
\rowcolor{gray!15}GPT-4o~\citep{hurst2024gpt} & 71.9 & - & 73.75 & 32 \\
\rowcolor{gray!15}Gemini-1.5-Pro~\citep{team2024gemini} & 75.0 & - & 67.1 & 32 \\
\midrule
\rowcolor{ModelPurple}\multicolumn{5}{l}{\textit{Open-Source Models}} \\
\midrule
VideoChat2~\citep{li2024mvbench} & 39.5 & 51.1 & 48.8 & 32 \\
LLaMA-VID~\citep{li2024llama} & - & 41.9 & 45.6 & 32 \\
VideoLLaVA~\citep{lin2024video} & - & 43.0 & 49.77 & 32 \\
ST-LLM~\citep{liu2024st} & 37.9 & 54.85 & - & 32 \\
ShareGPT4Video~\citep{chen2024sharegpt4video} & 39.9 & 51.2 & - & 32 \\
LLaVA-Next-Video~\citep{zhang2024llavanextvideo} & 60.2 & - & 66.74 & 32 \\
PLLaVA~\citep{xu2024pllava} & - & 46.6 & - & 32 \\
LongVA~\citep{zhang2024long} & 52.6 & - & 56.9 & 32 \\
VideoLLaMA2~\citep{cheng2024videollama} & 47.9 & 54.6 & - & 32 \\
Video-CCAM~\citep{fei2024video} & - & \textbf{64.6} & - & 32 \\
Long-LLaVA~\citep{wang2024longllava} & 52.9 & - & - & 32 \\
Kangaroo~\citep{liu2026kangaroo} & 56.0 & 61.1 & 62.5 & 32 \\

\rowcolor{ToolThinkBar}LongVT-RFT (LongVT, think with tools) \citep{yang2025longvt} & 59.7 & 53.9 & 68.3 & 32 \\
\rowcolor{VideoThinkBar}Video-Thinker-7B (think with video) \citep{wang2025video} & 60.0 & 62.4 & 67.8 & 32 \\

Qwen2.5-VL-7B-Instruct(CoT)~\citep{Bai2025Qwen25VLTR} & 60.8 & 42.4 & 72.8 & 32 \\
Qwen2.5-VL-7B-SFT & 55.4 & 60.5 & 69.9 & 32 \\

Video-R1-7B-wo-image~\citep{feng2026video} & 53.8 & 60.9 & 69.8 & 32 \\
Video-R1-7B~\citep{feng2026video} & 57.4 & 62.7 & 72.6 & 32 \\
\rowcolor{ModelGreen}\storms & \textbf{61.0} & 61.1 & \textbf{74.3} & 32 \\
\bottomrule
\end{tabular}%
}
\end{table*}

\begin{table*}[t]\small
\centering
\caption{Results on reasoning-heavy video benchmarks. The orange/blue bars indicate think-with-tools (LongVT) and think-with-video (Video-Thinker) with 32 max frames for a single round. Except for those two methods, all experiments use uniform sampling with 32 frames.}
\label{tab:reasoning_benchmarks}
\setlength{\tabcolsep}{4pt}
\resizebox{\textwidth}{!}{%
\begin{tabular}{lcccc}
\toprule
\textbf{Model} & \textbf{VideoEspresso} & \textbf{Video-Holmes} & \textbf{MMVU} & \textbf{Frames} \\
\midrule
\rowcolor{VideoThinkBar}\multicolumn{5}{l}{\textit{Proprietary Models}} \\
\midrule
\rowcolor{gray!15}GPT-4o~\citep{hurst2024gpt} & - & 42.0 & 75.4 & 32 \\
\rowcolor{gray!15}Gemini-1.5-Pro~\citep{team2024gemini} & 44.2 & 45.7 & - & 32 \\
\midrule
\rowcolor{ModelPurple}\multicolumn{5}{l}{\textit{Open-Source Models}} \\
\midrule
\rowcolor{ToolThinkBar}LongVT-RFT (LongVT, think with tools) \citep{yang2025longvt} & 52.7 & 35.5 & 61.4 & 32 \\
\rowcolor{VideoThinkBar}Video-Thinker-7B (think with video) \citep{wang2025video} & 56.4 & 43.1 & 63.4 & 32 \\

Qwen2.5-VL-7B-SFT & 49.6 & 31.4 & 62.1 & 32 \\
Qwen2.5-VL-7B-Instruct(CoT)~\citep{Bai2025Qwen25VLTR}  & 45 & 27.8 & 60.0 & 32 \\

Video-R1-7B~\citep{feng2026video} & - & 36.5 & 64.2 & 32 \\
\rowcolor{ModelGreen}\storms & \textbf{58.7} & \textbf{37.8} & \textbf{65.9} & 32 \\
\bottomrule
\end{tabular}%
}
\end{table*}

\begin{table*}[t]\small
\centering
\caption{Results on a smaller backbone, showing that \storms remains effective across model sizes.}
\label{tab:qwen25vl3b_placeholder}
\setlength{\tabcolsep}{5pt}
\begin{tabular}{lccc}
\toprule
\textbf{Model} & \textbf{VideoMME} & \textbf{TempCompass} & \textbf{MMVU} \\
\midrule
Qwen2.5-VL-3B-Instruct & 50.9 & 55.0 & 42.0 \\
\rowcolor{ModelGreen}\textbf{Qwen2.5-VL-3B + \storms} & \textbf{55.4} & \textbf{67.5} & \textbf{58.5} \\
\bottomrule
\end{tabular}
\end{table*}

\subsection{Benchmarks}
We evaluate our method on six benchmarks that cover both general video understanding and video reasoning: \textbf{VideoMME}~\citep{fu2026mme} contains 900 videos from six domains and approximately 2{,}700 multiple-choice questions over videos ranging from 11 seconds to 1 hour; \textbf{MVBench}~\citep{li2024mvbench} spans around 20 challenging tasks requiring temporal understanding and multimodal reasoning; \textbf{TempCompass}~\citep{liu2024tempcompass} focuses on temporal perception through questions about speed, direction, order, and change; \textbf{Video-Holmes}~\citep{cheng2025video} comprises 1{,}837 questions from 270 suspense short films and emphasizes clue-based multi-step reasoning; \textbf{VideoEspresso}~\citep{han2025videoespresso} is a large-scale chain-of-thought-based VideoQA benchmark built with semantic core-frame selection across 14 task types; and \textbf{MMVU}~\citep{zhao2025mmvu} evaluates expert-level multi-discipline video understanding, requiring models to combine detailed visual evidence with higher-level reasoning across diverse domains.

\subsection{Experiment Setup}
\noindent\textbf{Implementation details}. 
\storms is built upon Qwen2.5-VL-7B-Instruct and trained in two stages.
In Stage~I, we jointly optimize latent-state alignment and answer-token cross-entropy, with a balancing coefficient set to 0.1.
In Stage~II, we keep the latent simulation slots but remove direct latent supervision, and optimize only answer-token cross-entropy.
Each stage is trained for one epoch with a learning rate of $1\times10^{-5}$ and a batch size of 1.
To mitigate instability from variable latent lengths, we set the latent token budget to 8. All experiments are conducted on four NVIDIA GB200 GPUs.

\noindent\textbf{Datasets}.
For training, we use the video-based portion of the dataset introduced in~\citep{feng2026video}, which focuses on everyday scenarios and temporal reasoning. We exclude image-only samples for efficiency. Following the data construction pipeline in Section~\ref{data_pipeline}, we extract question-relevant keyframes with GPT-4o and generate 5K short thought videos conditioned on the keyframes and teacher CoT prompts. The resulting keyframe, thought-video, and answer annotations are used for the two-stage training.

\subsection{Results and Analysis}
\paragraph{Performance on general video benchmarks.}
We evaluate \storms on a suite of general video benchmarks, including VideoMME, MVBench, and TempCompass (Table~\ref{tab:general_benchmarks}). \storms consistently outperforms its standard supervised fine-tuned (SFT) counterpart, Qwen2.5-VL-7B-SFT, which is trained on the same dataset as \storms but without thought-video supervision. These improvements across spatial-temporal tasks suggest that the latent simulation objective provides benefits beyond standard supervised fine-tuning. Beyond the SFT baseline, \storms also matches or exceeds specialized models such as LongVT and Video-Thinker, which rely on expensive test-time compute pipelines. This comparison suggests that latent-space reasoning can provide similar benefits to explicit external reasoning while avoiding the same inference bottlenecks.

\textbf{Generalizability to smaller backbones.} To examine whether these benefits are tied to a single model scale, we also validate \storms on the smaller Qwen2.5-VL-3B backbone. As shown in Table~\ref{tab:qwen25vl3b_placeholder}, \storms improves baseline performance across VideoMME, TempCompass, and MMVU, supporting the framework's generalizability.

\paragraph{Performance on reasoning-focused video benchmarks.}

Similar improvements emerge on reasoning-focused benchmarks, including VideoEspresso, Video-Holmes, and MMVU (Table~\ref{tab:reasoning_benchmarks}). \storms consistently outperforms standard SFT baselines, suggesting that internalized visual reasoning can help with complex, multi-step logic. Even when evaluated against state-of-the-art models that rely on explicit video generation or complex external toolchains, \storms exhibits competitive performance. These results suggest that implicit latent simulation can serve as an efficient alternative to resource-heavy pipelines, supporting strong video understanding without adding substantial inference-time complexity.

\begin{table}[t]
    \centering
    \caption{Latency analysis on MMVU using provided runtime statistics. \storms latency is converted from case-study throughput (2.14 it/s), corresponding to about 0.47 s/item. Lower is better.}
    \label{tab:latency_mmvu}
    \small
    \setlength{\tabcolsep}{5pt}
    \renewcommand{\arraystretch}{1.12}
    \begin{tabular}{lcc}
        \toprule
        \textbf{Model} & \textbf{Throughput} & \textbf{Latency (s/item)} \\
        \midrule
        LongVT & 0.065 it/s & 15.44 \\
        Video-Thinker & 0.058 it/s & 17.20 \\
        \rowcolor{ModelGreen}
        \storms & 2.14 it/s & \textbf{0.47} \\
        \bottomrule
    \end{tabular}
\end{table}

\paragraph{Latency analysis.}
We further compare runtime efficiency on MMVU in Table~\ref{tab:latency_mmvu}. LongVT and Video-Thinker require 15.44 and 17.20 s/item, corresponding to only 0.065 and 0.058 items/s, respectively. In contrast, \storms processes 2.14 items/s, or approximately 0.47 s/item. This efficiency gap suggests that internalizing the reasoning process can make \storms more efficient than methods that rely on costly test-time tool calls or repeated frame reinsertion. During inference, \storms carries out intermediate reasoning through a compact set of latent tokens, keeping the additional reasoning cost bounded and predictable. With this training design, the expensive visual reasoning signal is introduced through thought-video supervision during training, while inference remains a compact single-model decoding process. This reduces test-time overhead compared with methods that explicitly revisit visual inputs or invoke external reasoning modules.

\begin{figure*}[t]
    \centering
    \includegraphics[width=0.88\textwidth]{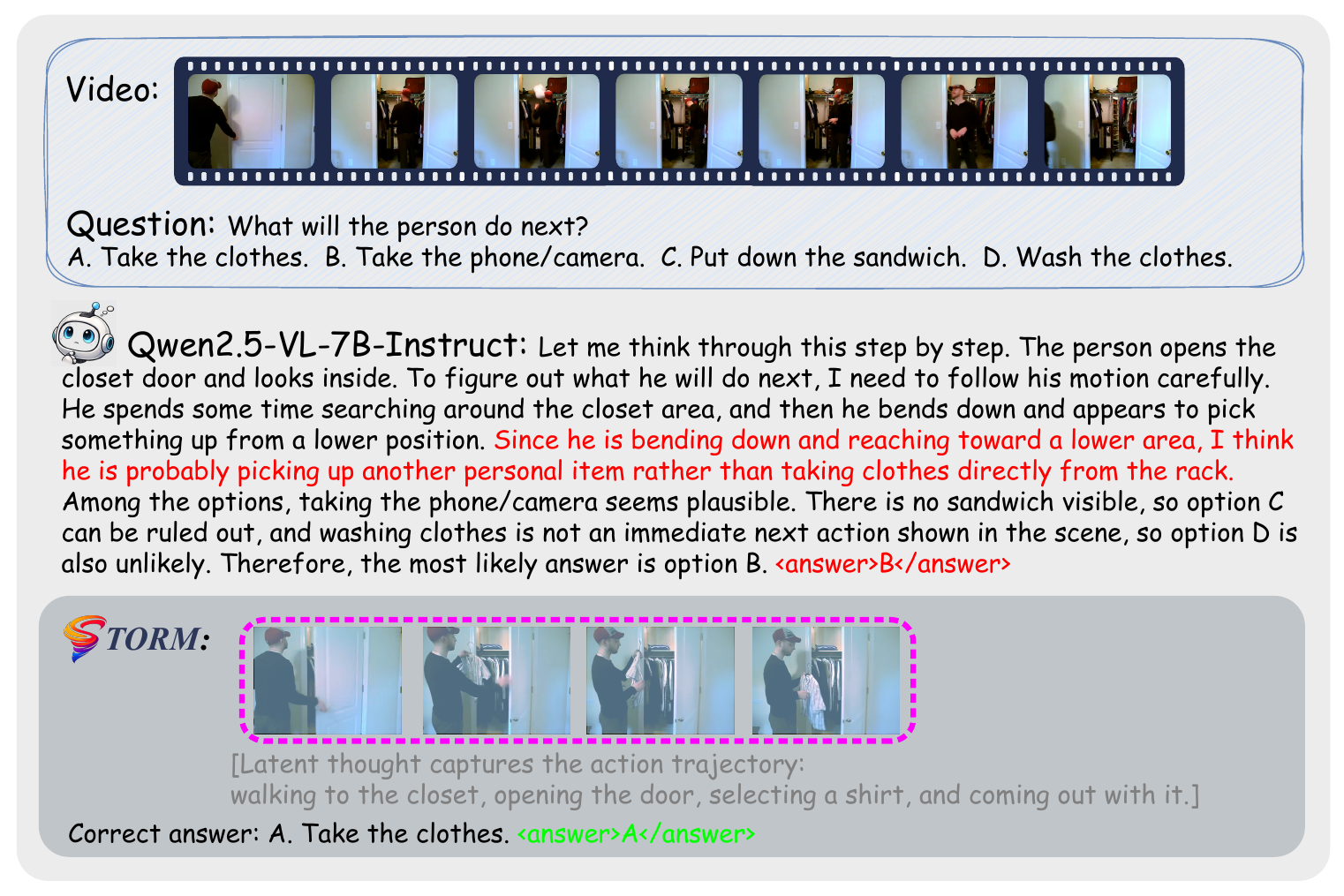}
    \caption{Case study comparing textual reasoning with \storms latent reasoning on future-action prediction. \storms captures the action trajectory in latent space and recovers the correct answer.}
    \label{fig:case_study}
\end{figure*}

\subsection{Ablations}

\paragraph{Effects of different latent sizes.} To study how the latent-slot budget affects performance, we vary the number of latent tokens used to compress thought-video features. As shown in Table~\ref{tab:dynamic_compression}, very small latent budgets provide limited capacity for representing temporal-visual evidence, while overly large budgets can dilute the supervision signal and introduce redundant visual tokens. The best performance is achieved with $K=8$, which provides a compact but sufficiently expressive latent trajectory. In contrast, increasing the latent size to 16 or removing adaptive pooling entirely does not improve performance, suggesting that preserving more visual tokens is not necessarily beneficial for internalized reasoning. We therefore use $K=8$ in all main experiments.

\begin{table}[t]
    \centering
    \begin{minipage}[t]{0.49\textwidth}
        \centering
        \caption{Evaluation across latent-token sizes. A moderate budget of 8 latent tokens achieves the best overall balance across all three benchmarks, while larger budgets degrade performance, suggesting that compact latent trajectories provide more effective supervision than retaining redundant visual tokens.}
        \label{tab:dynamic_compression}
        \scriptsize
        \setlength{\tabcolsep}{3pt}
        \renewcommand{\arraystretch}{1.05}
        \resizebox{\linewidth}{!}{%
        \begin{tabular}{lccc}
            \toprule
            \textbf{Latent size} & \textbf{VideoMME} & \textbf{TempCompass} & \textbf{MVBench} \\
            \midrule
            4                & 58.9 & 71.1 & 55.0 \\
            \rowcolor{ModelPurple}
            \textbf{8}                & \textbf{61.0} & \textbf{74.3} & \textbf{61.1} \\
            16               & 52.2 & 67.3 & 55.5 \\
            256 (no average) & 52.9 & 68.7 & 54.7 \\
            \bottomrule
        \end{tabular}%
        }
    \end{minipage}
    \hfill
    \begin{minipage}[t]{0.49\textwidth}
        \centering
        \caption{Performance by two-stage training configuration. The full \storms pipeline consistently outperforms either stage alone, indicating that thought-video grounded latent alignment in Stage~I and answer-focused consolidation in Stage~II are complementary.}
        \label{tab:stages}
        \scriptsize
        \setlength{\tabcolsep}{3pt}
        \renewcommand{\arraystretch}{1.05}
        \resizebox{\linewidth}{!}{%
        \begin{tabular}{lccc}
            \toprule
            \textbf{Setting} & \textbf{VideoMME} & \textbf{TempCompass} & \textbf{MVBench} \\
            \midrule
            Stage~I only & 58.8 & 72.0 & 56.0 \\
            Stage~II only & 56.6 & 72.6 & 56.8 \\
            \rowcolor{ModelPurple}
            \textbf{Full \storms (Stage~I + Stage~II)} & \textbf{61.0} & \textbf{74.3} & \textbf{61.1} \\
            \bottomrule
        \end{tabular}%
        }
    \end{minipage}
\end{table}

\paragraph{Effects of different training stages.} Table~\ref{tab:stages} compares using only Stage~I, using only Stage~II, and the full two-stage \storms pipeline. Both single-stage variants underperform the full model: Stage~I learns grounded latent dynamics from thought-video supervision but lacks large-scale answer alignment, while Stage~II benefits from answer-token supervision but lacks latent-space grounding. The best performance therefore requires both thought-video-grounded latent supervision in Stage~I and answer-focused consolidation in Stage~II.

\begin{figure}[H]
    \centering
    \includegraphics[width=0.82\textwidth]{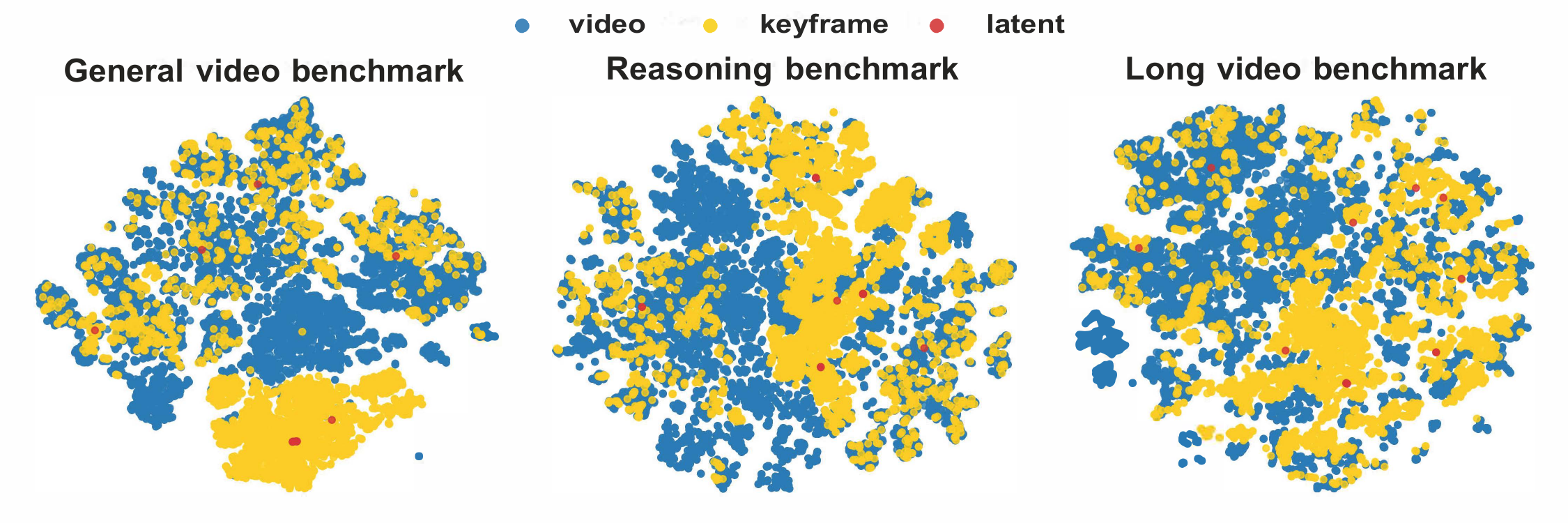}
    \caption{t-SNE plot of latent-token, keyframe, and video-frame embeddings.}
    \vspace{-5pt}
    \label{fig:tsne}
\end{figure}

\paragraph{Latent token analysis.} We sample 20 examples across three benchmark categories and visualize video-frame, keyframe, and latent-token embeddings after PCA and t-SNE projection. As shown in Figure~\ref{fig:tsne}, keyframe embeddings form sub-clusters within the broader video-frame distribution, and the sparse latent-token embeddings consistently fall inside or adjacent to the corresponding keyframe regions. This indicates that \storms learns informative latent representations aligned with true keyframe semantics, enabling more effective video reasoning.

\FloatBarrier

\section{Conclusion}
In this paper, we present \storms, a framework for internalized spatial-temporal reasoning in video-language models. Instead of relying on explicit textual CoT, repeated frame reinsertion, or external visual tools, \storms captures key temporal-visual evidence through bounded latent trajectories. By using generated thought videos only during training, \storms moves visual reasoning from inference-time tool orchestration toward an internalized latent process, enabling strong benchmark performance while reducing test-time overhead. We hope this work provides a practical step toward efficient native visual reasoning in video-language models.



\newpage
\bibliographystyle{plainnat}
\bibliography{neurips_2026}



\newpage
\appendix

\section{Additional Analysis of Latent Slot Representations}
\label{app:latent_slot_analysis}

\subsection{Same-video retrieval probe}
\label{app:same_video_retrieval}

To further examine whether the learned latent states encode video-specific temporal information, we conduct a same-video retrieval probe. For each query sample, we use its learned latent representation to retrieve other samples and measure whether the nearest retrieved examples originate from the same source video. 

\begin{figure}[H]
    \centering
    \includegraphics[width=0.68\linewidth]{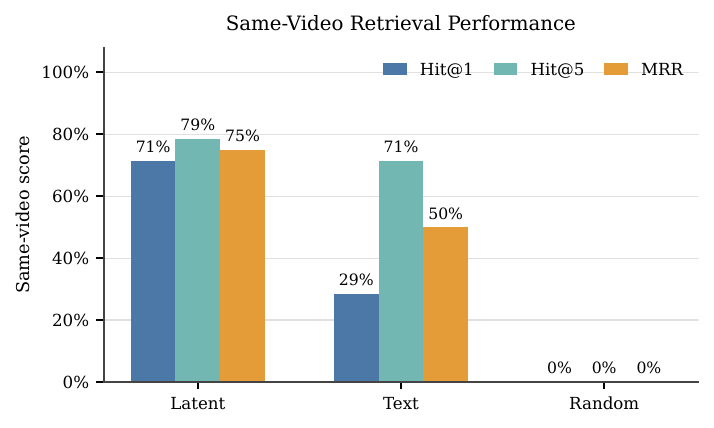}
    \caption{\textbf{Same-video retrieval performance.} We compare latent representations, textual representations, and random retrieval using Hit@1, Hit@5, and MRR. Latent representations achieve the strongest same-video retrieval performance, indicating that the latent states preserve video-specific information.}
    \label{fig:retrieval_summary}
\end{figure}

Figure~\ref{fig:retrieval_summary} compares retrieval using latent representations, textual representations, and a random baseline. Latent representations achieve substantially stronger same-video retrieval performance, reaching 71\% Hit@1, 79\% Hit@5, and 75\% MRR. In comparison, textual representations achieve 29\% Hit@1, 71\% Hit@5, and 50\% MRR, while random retrieval gives 0\% under all three metrics. This suggests that the learned latent states retain video-specific information more reliably than text-only representations, especially at the top retrieval rank. The large gap in Hit@1 indicates that latent states are more effective at identifying the most relevant same-video neighbor, rather than only recovering a same-video example somewhere in the top-ranked set.

\subsection{Per-query retrieval behavior}
\label{app:per_query_retrieval}

\begin{figure}[H]
    \centering
    \IfFileExists{query_retrieval_comparison.pdf}{%
        \includegraphics[width=0.82\linewidth]{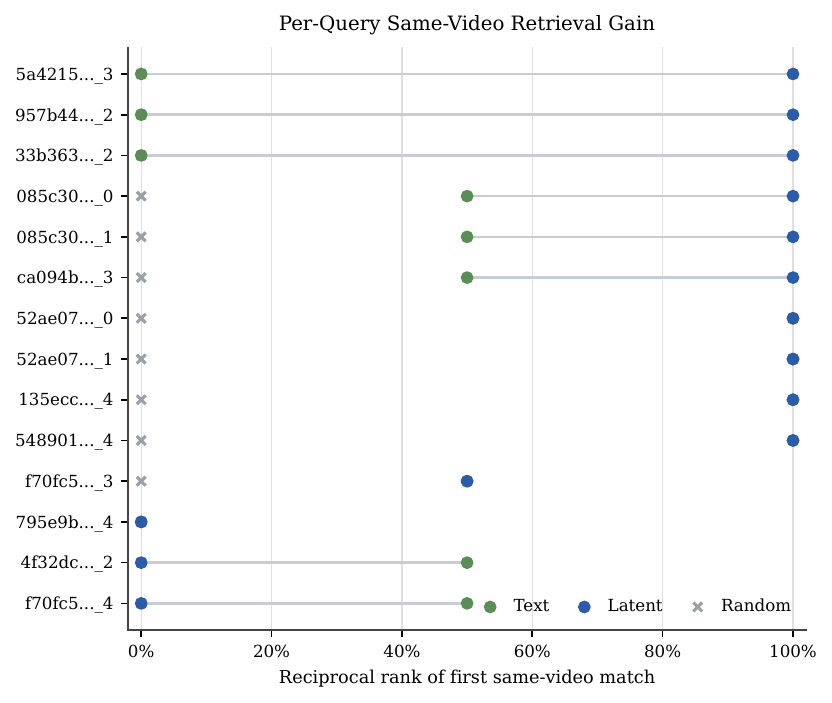}%
    }{%
        \fbox{%
            \parbox{0.76\linewidth}{%
                \centering
                Per-query retrieval figure unavailable in the current workspace.\\
                Replace this placeholder with \texttt{query\_retrieval\_comparison.pdf} before submission.%
            }%
        }%
    }
    \caption{\textbf{Per-query same-video retrieval gain.} Each row corresponds to a query example, and the x-axis shows the reciprocal rank of the first retrieved same-video match. Latent representations more frequently retrieve same-video examples at high rank, especially at rank 1.}
    \label{fig:query_retrieval_comparison}
\end{figure}

We additionally visualize the reciprocal rank of the first same-video match for each query in Figure~\ref{fig:query_retrieval_comparison}. This per-query view shows that the improvement from latent representations is not caused by a single outlier. For several queries, latent representations retrieve a same-video match at rank 1, while text representations either fail or retrieve the match at a much lower rank. In particular, multiple examples achieve a reciprocal rank of 1.0 under the latent representation, corresponding to perfect top-1 same-video retrieval. Text representations occasionally retrieve same-video examples, but they more often assign them lower ranks. Random retrieval remains near zero across nearly all queries.

This result provides complementary evidence to the aggregate metrics in Figure~\ref{fig:retrieval_summary}. While the aggregate results show that latent representations improve overall retrieval quality, the per-query comparison shows that the gain is distributed across many samples. This supports the interpretation that \storms latent states encode temporally grounded video information rather than merely memorizing answer labels or exploiting a small number of easy examples.

\subsection{Latent slot ablation}
\label{app:slot_ablation}

\begin{figure}[H]
   \centering
   \includegraphics[width=0.72\linewidth]{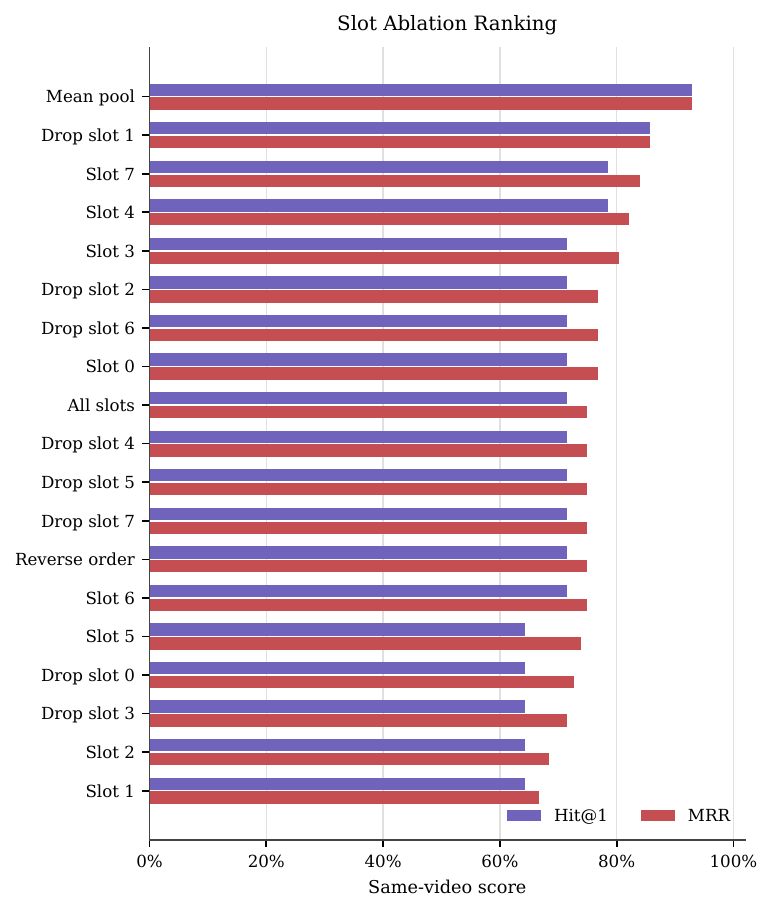}
   \caption{\textbf{Latent slot ablation ranking.} We compare same-video retrieval performance under different slot selection, dropping, reversal, and pooling strategies. Mean pooling performs best, indicating that useful video-specific information is distributed across the latent trajectory.}
   \label{fig:slot_ablation}
\end{figure}

We further analyze which latent slots contribute most to same-video retrieval. Figure~\ref{fig:slot_ablation} ranks different slot aggregation and ablation strategies according to same-video retrieval performance. Mean pooling over latent slots achieves the strongest performance among the tested variants, suggesting that video-specific information is distributed across multiple latent states rather than concentrated in a single slot. Dropping individual slots generally reduces performance only moderately, which indicates that the latent representation is relatively robust to the removal of any one slot.

Among single-slot variants, later slots such as Slot 7 and Slot 4 perform strongly, while earlier slots such as Slot 1 and Slot 2 are weaker. This trend suggests that later latent states may accumulate more complete temporal evidence after several rounds of latent rollout. However, no single slot consistently outperforms mean pooling, supporting the design choice of treating the latent rollout as a short trajectory rather than relying on a single hidden state. The reverse-order variant performs similarly to several slot-dropping variants, suggesting that the learned representation is not purely order-agnostic but also not catastrophically sensitive to slot ordering under this diagnostic.

Overall, the slot ablation supports two conclusions. First, the latent rollout contains meaningful video-level information that can be recovered through retrieval. Second, this information is distributed across the latent trajectory, with mean pooling providing a stable summary. These findings are consistent with the main design of \storms, where multiple latent slots jointly form a compact internal simulation before answer generation.

\section{Prompt Templates for Data Construction}
\label{app:prompt_templates}

This appendix provides the prompt templates used in our data construction pipeline. 
Appendix~\ref{app:keyframe_selection_prompt} presents the prompt for selecting question-relevant keyframes from candidate frames, and Appendix~\ref{app:thought_video_generation_prompt} presents the prompt for generating thought videos from the selected keyframe and teacher-generated reasoning plan.

\subsection{Keyframe Selection Prompt}
\label{app:keyframe_selection_prompt}

\begin{figure}[H]
    \centering
    \includegraphics[width=0.95\linewidth]{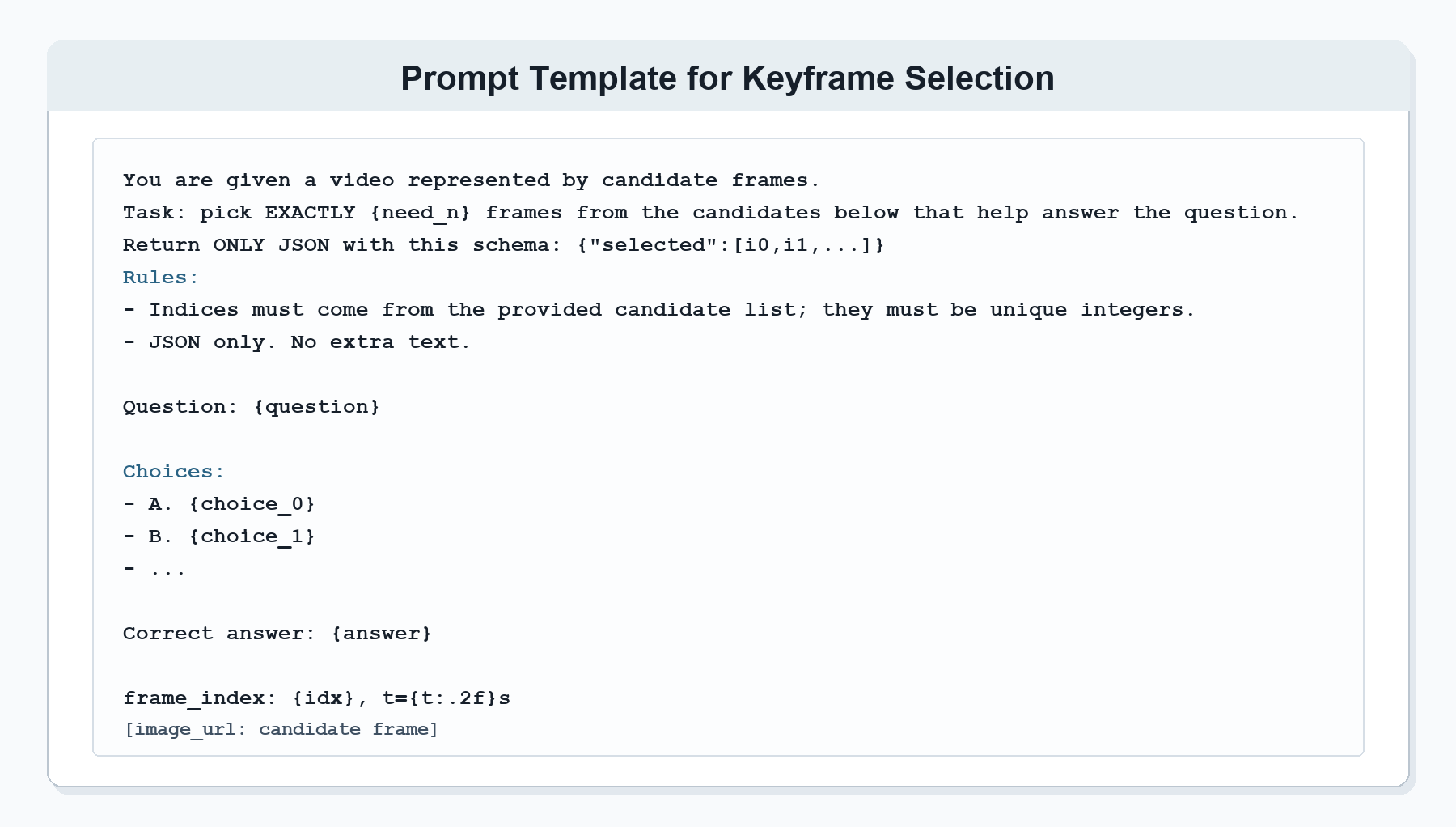}
    \caption{\textbf{Prompt template for keyframe selection.} The prompt asks GPT-4o to select exactly the required number of question-relevant frames from a candidate-frame list and return only the selected frame indices in JSON format.}
    \label{fig:keyframe_selection_prompt}
\end{figure}

Figure~\ref{fig:keyframe_selection_prompt} shows the prompt used to select question-relevant keyframes from candidate frames. 
Given a question, optional answer choices, the correct answer, and a list of candidate frames with timestamps, GPT-4o is instructed to return only the selected frame indices in JSON format.

\subsection{Thought-Video Generation Prompt}
\label{app:thought_video_generation_prompt}

\begin{figure}[H]
    \centering
    \includegraphics[width=0.95\linewidth]{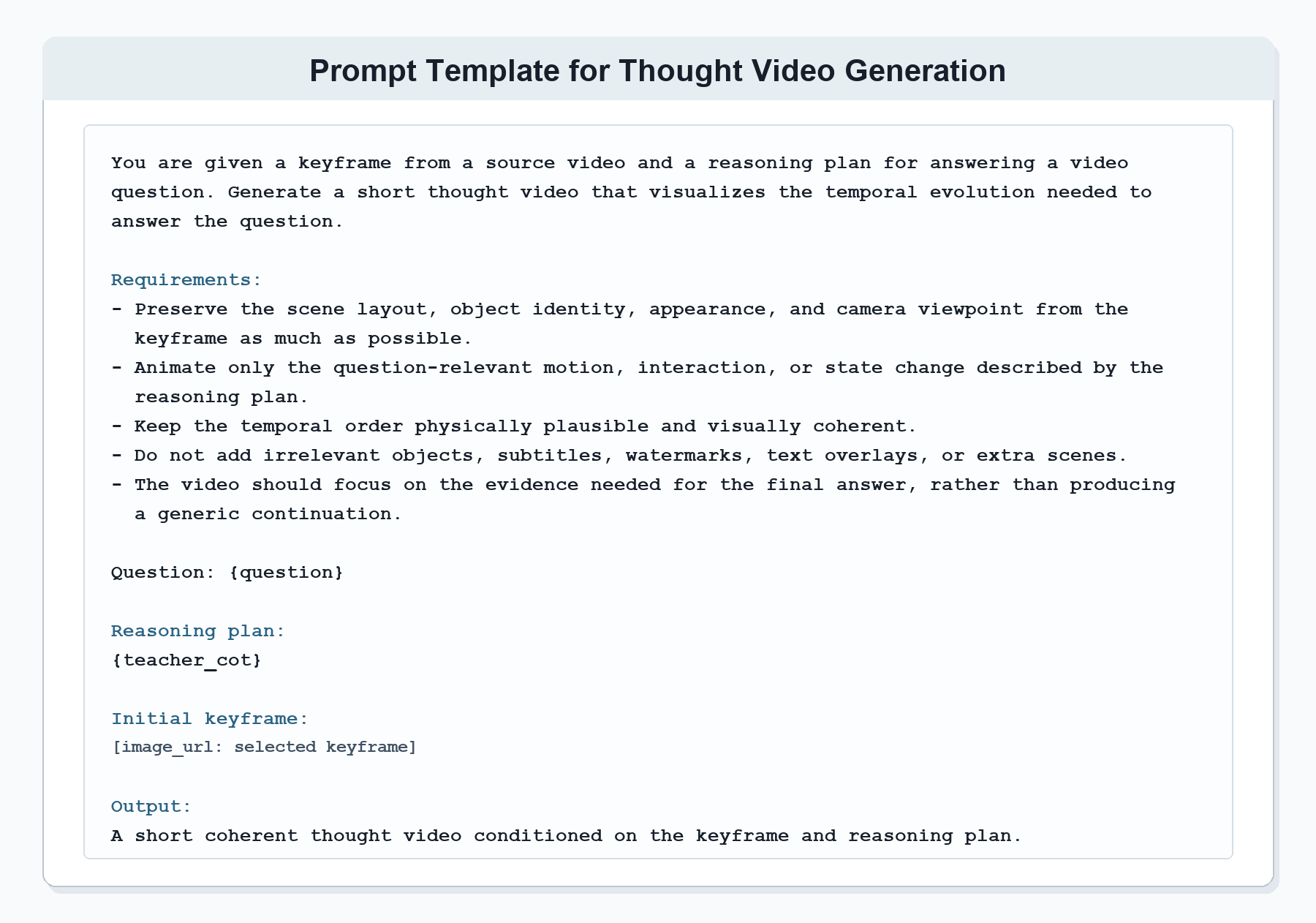}
    \caption{\textbf{Prompt template for thought-video generation.} The prompt conditions the video generator on the selected keyframe, the question, and the teacher-generated reasoning plan, encouraging a short coherent video that visualizes the temporal evidence needed to answer the question.}
    \label{fig:thought_video_generation_prompt}
\end{figure}

Figure~\ref{fig:thought_video_generation_prompt} shows the prompt used to generate thought videos. 
The video generator is conditioned on the selected keyframe, the question, and the teacher-generated reasoning plan. 
The generated thought video is used only as training-time supervision for latent-state alignment.

\section{Limitations and Future Work}
\label{app:limitations}

Although \storms provides an efficient framework for internalized spatial-temporal reasoning, several limitations remain. First, it relies on generated thought-video supervision during training, whose quality depends on keyframe extraction, teacher CoT prompts, and the video generation model; noisy keyframes or inaccurate generated motion may lead to imperfect latent targets. Second, our evaluation mainly focuses on multiple-choice video QA benchmarks and Qwen2.5-VL backbones, so broader validation on open-ended video reasoning, dense temporal grounding, embodied tasks, and other LVLM architectures remains future work. Finally, while \storms reduces inference-time cost, it introduces additional training-time cost for constructing thought-video supervision. Future work will explore stronger filtering, multi-keyframe conditioning, architecture transfer, and lighter supervision strategies such as keyframe-only targets or self-distilled latent targets.

\newpage

\end{document}